f
\documentclass{article}
\usepackage{spconf,amsmath,graphicx,multirow,array,tikz,subfig,bm,amsfonts}
\newcommand{\PreserveBackslash}[1]{\let\temp=\\#1\let\\=\temp}
\newcolumntype{C}[1]{>{\PreserveBackslash\centering}p{#1}}
\newcolumntype{R}[1]{>{\PreserveBackslash\raggedleft}p{#1}}
\newcolumntype{L}[1]{>{\PreserveBackslash\raggedright}p{#1}}

\title{STATE-OF-THE-ART SPEECH RECOGNITION\\USING MULTI-STREAM SELF-ATTENTION WITH DILATED 1D CONVOLUTIONS}
\name{Kyu J. Han, Ramon Prieto, Kaixing Wu, Tao Ma}
\address{ASAPP Inc.\\Mountain View, CA, USA\\
\texttt{\{khan,rprieto,kwu,tma\}@asapp.com}}
\begin{document}
%
\maketitle
\begin{abstract}
Self-attention has been a huge success for many downstream tasks in NLP, which led to exploration of applying self-attention to speech problems as well. The efficacy of self-attention in speech applications, however, seems not fully blown yet since it is challenging to handle highly correlated speech frames in the context of self-attention. In this paper we propose a new neural network model architecture, namely \textit{multi-stream self-attention}, to address the issue thus make the self-attention mechanism more effective for speech recognition. The proposed model architecture consists of parallel streams of self-attention encoders, and each stream has layers of 1D convolutions with dilated kernels whose dilation rates are unique given stream, followed by a self-attention layer. The self-attention mechanism in each stream pays attention to only one resolution of input speech frames and the attentive computation can be more efficient. In a later stage, outputs from all the streams are concatenated then linearly projected to the final embedding. By stacking the proposed multi-stream self-attention encoder blocks and rescoring the resultant lattices with neural network language models, we achieve the word error rate of 2.2\% on the test-clean dataset of the LibriSpeech corpus, the best number reported thus far on the dataset.
\end{abstract}
\section{Introduction}
\label{sec:intro}

Self-attention is the core component of the neural network architectures recently proposed in NLP \cite{vaswani,devlin,Yang2019XLNetGA} to achieve the state-of-the art performances in a number of downstream tasks. Transformer \cite{vaswani} successfully replaced recurrent neural networks such as LSTMs with sinusoidal positional encoding and the self-attention mechanism to be context-aware on input word embeddings. BERT \cite{devlin} took the benefit from the success of Transformer to extend it to the autoencoding based pretraining model, which can be fine-tuned to reach the state-of-the-art performances for various downstream tasks. XLNet \cite{Yang2019XLNetGA}, as the very latest state-of-the-art pretraining model, outperformed BERT in a number of downstream tasks from question answering to document ranking, thanks to model training with targets being aware and relative positional encoding like its ancestor of Transformer-XL \cite{Dai2019TransformerXLAL}.

With the huge success in NLP, self-attention has been actively investigated for speech recognition as well \cite{povey18,Chiu18,Sperber18,Salazar19,Dong2018SpeechTransformerAN,zhou18}. In \cite{povey18}, time-restricted self-attention was introduced with a one-hot vector representation being exploited as relative positional encoding for given restricted contexts of speech frames. In \cite{Chiu18}, the well-known Listen, Attend and Spell (LAS) ASR model \cite{chan} employed the multi-head approach to the attention mechanism to further improve its already state-of-the-art accuracy on the large-scale voice search data. In \cite{Sperber18}, several approaches were explored for better application of self-attention to speech recognition in the LAS framework, e.g., speech frame handling strategies or attention biasing to restrict the locality of the self-attention mechanism. In \cite{Salazar19}, the CTC loss \cite{graves06} was applied to optimize the Transformer encoder structure for ASR. In \cite{Dong2018SpeechTransformerAN,zhou18}, the entire encoder-decoder structure of the original Transformer \cite{vaswani} was examined in the context of Mandarin Chinese speech recognition tasks. 

The challenge in terms of applying self-attention to speech recognition is that individual speech frames are not like lexical units such as words. Speech frames do not convey distinct meanings or perform unique functions, which makes it hard for the self-attention mechanism to compute proper attentive weights on speech frames. Considering that adjacent speech frames could form a chunk to represent more meaningful units like phonemes, some sort of pre-processing mechanisms such as convolutions to capture an embedding for a group of nearby speech frames would be helpful for self-attention. In addition, a multi-resolution approach could be beneficial as well since boundaries for such meaningful chunks of speech frames are dependent of many factors, e.g., the type of phonemes (vowel vs. consonant) and the way they are pronounced, affected by gender, speaker, co-articulation and so on. Based on this reasoning, in this paper, we propose a new neural network model architecture for better self-attention, namely \textit{multi-stream self-attention}.  The proposed architecture consists of parallel streams of self-attention. In each stream, input speech frames are processed with a distinct resolution by multiple layers of 1D convolutions with a unique dilation rate, and the convoluted embeddings are fed to a subsequent multi-head self-attention layer. In a later stage, the attentive embeddings from all the streams are concatenated then linearly projected to the final embedding. We achieve the state-of-the-art performances on the LibriSpeech corpus \cite{panayotov2015librispeech} by stacking up these multi-stream self-attention blocks and rescoring the resultant lattices with powerful neural language models. Our WERs on the dev-clean and test-clean sets of 1.8\% and 2.2\%, respectively, are the best reported numbers thus far on the datasets as far as we know. 

\begin{figure}[t]
  \centering
  \centerline{\includegraphics[width=11cm]{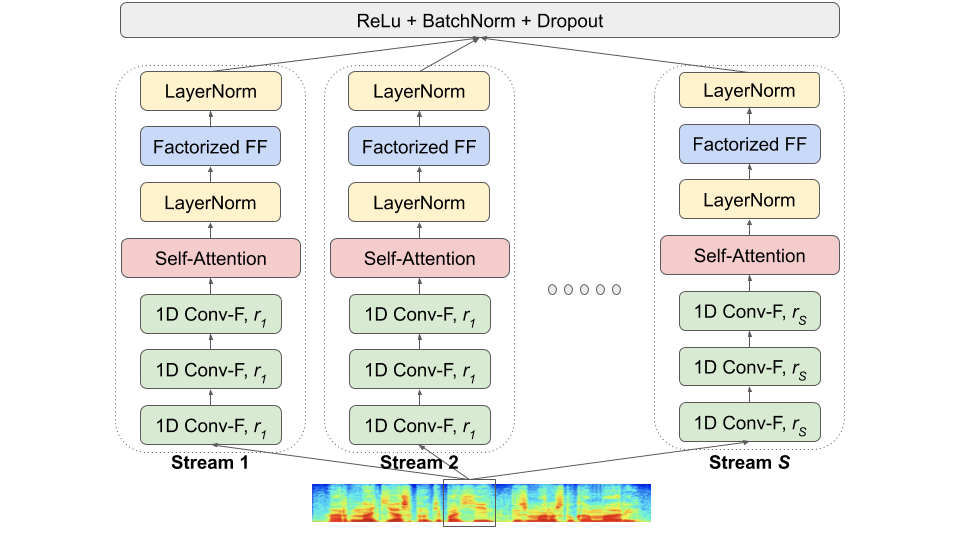}}
  \caption{Multi-stream self-attention block. $r_{S}$: dilation rate for the 1D convolutions in the stream $S$.}
\end{figure}

This paper is organized in the following structure. In Section 2, we provide the details of the proposed multi-stream self-attention model architecture. In Section 3, we present the experimental setups and discuss the validity of the proposed ideas using the ablation tests. In addition, we compare our ASR system based on the multi-stream self-attention models with the other state-of-the-art systems. In Section 4, we conclude the paper with the summary of our contributions as well as the future directions.

\section{MULTI-STREAM SELF-ATTENTION}
\label{sec:format}

Fig. 1 shows the proposed multi-stream self-attention block, each stream of which consists of multiple layers of 1D convolutions, one layer of multi-head self-attention, and a feed forward layer sandwiched by two layer normalizations\footnote{Each layer normalization has a skip connection (not depicted in the figure) with the input of its previous layer.} \cite{ln}. The embeddings from all the streams are projected to the final embedding in a later stage. In this section, we detail each component of the multi-stream self-attention architecture. 

\begin{figure}[t]
  \centering
  \centerline{\includegraphics[width=8cm]{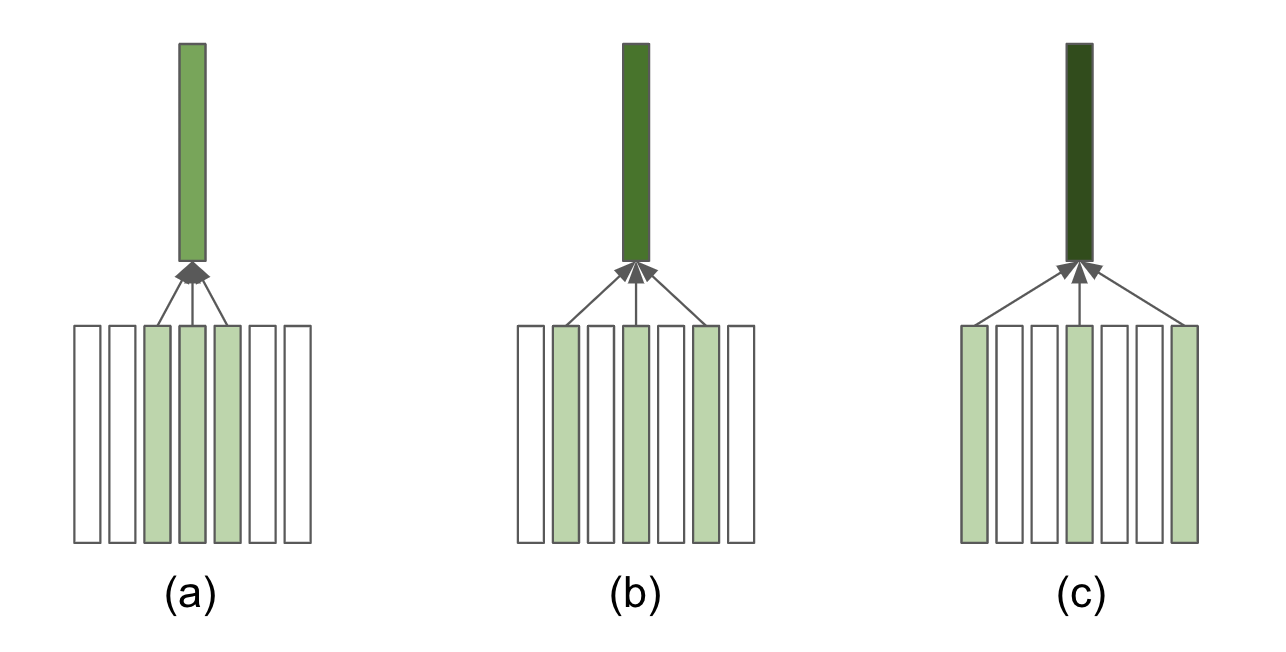}}
  \caption{Examples of the 1D convolutions with the $3 \times 1$ kernels and the different dilation rates ($r$). (a): $r = 1$, (b): $r = 2$, (c): $r = 3$.}
\end{figure}

\subsection{1D convolutions with factorization (1D Conv-F)}
Time delay neural networks (TDNNs) have been one of the most popular neural network models for speech recognition \cite{tdnn,tdnn15povey}. It was introduced to capture the long range temporal dependencies of acoustic events in speech signals \cite{tdnn} by exploiting a modular and incremental design from sub-components. The modified version was recently proposed for better efficiency using the layer-wise subsampling methods \cite{tdnn15povey}. 

A TDNN is basically a 1D convolution. We use this convolution layer and its kernels with various dilation rates to control the resolution of input speech frames being processed in the parallel streams. In each stream, layers of 1D convolutions with a a unique dilation rate process speech frames in the specified resolution. This can reduce burden on the self-attention mechanism to enable the attentive computation to focus on only one resolution of speech frames in the given stream. Examples of the 1D convolutions which the $3 \times 1$ kernels with the dilation rates of 1, 2, and 3 are applied to are shown in Fig. 2.

In order to make the convolution layers more efficient, we utilize the factorized TDNN \cite{povey18tdnnf}. Singular Value Decomposition (SVD) has been a popular choice to factorize a learned weight matrix into two low-rank factors and reduce the model complexity of neural networks \cite{xue13,rohit16,Tucker+2016}. The factorized TDNN or factorized 1D convolution  (1D Conv-F) layers also utilizes SVD to factorize a 1D convolution parameter matrix into two low-rank matrices. The kernel for each factorized 1D convolution is set to $2 \times 1$. One of the two factorized convolutions is constrained by the semi-orthogonal condition during training. Consider $\bm{U}$ as one of the factors in the original parameter matrix $\bm{W}$ after SVD. The semi-orthogonal constraint puts the condition to minimize a function $f$:  
\begin{equation}
    f = \texttt{Trace} \left( \bm{Q} \bm{Q} ^T \right)
\end{equation}
where $\bm{Q} = \bm{P} - \bm{I}$. $\bm{P}$ is defined by $\bm{U} \bm{U}^T$ and $\bm{I}$ is the identity matrix. This way of factorization with the semi-orthogonal constraint not only leads to less model complexity overall, but also results in even better modeling power. After the factorization, the rectified linear unit (ReLu) \cite{relu}, batch normalization \cite{bn}, and dropout \cite{dropout} are followed by a skip connection \cite{He16} between the scaled input embedding and the output of the dropout layer. The scale value is a hyper-parameter.

\subsection{Self-attention with factorized feed-forward (FF)}
In a given stream of $s$, we can formulate the time-restricted self-attention mechanism \cite{povey18} in a mathematical manner as follows. We define an input embedding matrix to the stream $s$, $\bm{X^s} \in \mathbb{R}^{N \times d_{model}}$, where $N$ is the total number of input embeddings restricted by the left and right context and $d_{model}$ is the dimension of embeddings used inside the self-attention mechanism. Note that downsampling is applied to the input embeddings and the sampling rate is matched to the specified dilation rate ($r_s$) of the 1D Conv-F layers in the stream. For the projected query, key and value matrices, $\bm{Q}_i^s, \bm{K}_i^s$ and $\bm{V}_i^s$ in the stream $s$, the output for the $i^{\textrm{th}}$ head is computed as follows:
\begin{equation}
 \textit{Head}_{i}^s  = \texttt{Softmax} \left( \frac{\bm{Q}_i^s {\bm{K}_{i}^s}^T}{\sqrt{d_k}} \right) \bm{V}_i^s
\end{equation}
where $\bm{Q}_i^s = \bm{X}^s \bm{W}_i^{s,Q}$ and $\bm{W}_i^{s,Q} \in \mathbb{R}^{d_{model} \times d_q}$, $\bm{K}_i^s = \bm{X}^s \bm{W}_i^{s,K} $ and $\bm{W}_i^{s,K} \in \mathbb{R}^{d_{model} \times d_k}$, $\bm{V}_i^s = \bm{X}^s \bm{W}_i^{s,K}$ and $\bm{W}_i^{s,V} \in \mathbb{R}^{d_{model} \times d_v}$, $d_q, d_k$, and $d_v$ are the dimensions of query, key and value embeddings, respectively. The multi-head outputs are concatenated and linearly projected, then layer normalization is applied to the projected embedding that is skip-connected with the input embedding:
\begin{align}
    \textit{MultiHeadProj}^s &= \texttt{Concat} \left( \textit{Head}_1^s, \dots, \textit{Head}_{n_h^s}^s \right) \bm{W}^{s,O} \\
  \textit{MidLayer}^s &= \texttt{LayerNorm} \left( \textit{MultiHeadProj}^s +  \bm{X}^s \right)
\end{align}
where $n_h^s$ is the number of heads in the stream $s$ and $\bm{W}_i^{s,O} \in \mathbb{R}^{\left( n_h \times d_v \right)  \times d_{model}}$. We set $n_h^s = n_h / S$, given $n_h$ is a fixed value for the total number of multi-heads across the self-attention components of the whole streams.

The multiple streams in the proposed architecture could increase the model complexity significantly as opposed to a single stream approach. To retain the model complexity to a reasonable level as well as avoid the loss of modeling power, we use the factorized feed forward networks with a bottleneck layer in between given stream. The semi-orthogonal constraint discussed in Section 2.1 is also applied here to one of the factorized matrices during training. After the skip connection, the encoder output in the stream $s$ can be written as below:
\begin{equation}
\textit{Factorized}^s  = \texttt{Factorized-FF} \left(  \textit{MidLayer}^s  \right) 
\end{equation}
\begin{equation}
\textit{Encoder}^s  = \texttt{LayerNorm} \left( \textit{Factorized}^s  + \textit{MidLayer}^s  \right)    
\end{equation}
Note that the dimension $d_{ff}$  of the embedding layer between the feed forward networks of the original Transformer encoder \cite{vaswani} is either 1,024 or 2,048. We could reduce the model complexity of the feed forward component by the factor of 8 or 16 if our choice of the bottleneck layer in the factorized version were 128. We can discuss how the factorized feed forward networks with this narrower bottleneck dimension are compared to the original version with no factorization but wider dimension in Section 3. 

\subsection{Final embedding}
The final embedding layer concatenates the encoder output from each stream and linearly projects the concatenated vector to the final embedding. The ReLu non-linear activation, batch normalization, and dropout follows before feeding out the final embedding as the output:
\begin{equation}
    \textit{MultiEncProj} = \texttt{Concat} \left( \textit{Encoder}^1, \dots, \textit{Encoder}^{S} \right) \bm{W}^O
\end{equation}
\begin{equation}
  \textit{Final} = \texttt{Dropout} \left( \texttt{BatchNorm} \left( \texttt{ReLu} \left( \textit{MultiEncProj} \right) \right) \right)
\end{equation}
where $\bm{W}^O \in \mathbb{R}^{\left( S \times d_{model} \right)  \times d_{model}}$.

\section{EXPERIMENTS AND RESULTS}
\subsection{Data and experimental setups}
For the experiments being discussed in the paper, we use the LibriSpeech corpus \cite{panayotov2015librispeech} as the main training and testing datasets. The LibriSpeech corpus is a collection of approximately 1,000hr audiobooks that are a part of the LibriVox project \cite{librivox14}. Most of the audiobooks come from the Project Gutenberg\footnote{http://www.gutenberg.org.}. The training data is split into 3 partitions of 100hr, 360hr, and 500hr sets while the dev and test data are split into the 'clean' and 'other' categories, respectively, depending upon how well or challening ASR systems would perform against. Each of the dev and test sets is around 5hr in audio length. This corpus also provides the n-gram language models and the corresponding texts\footnote{Available at http://www.openslr.org/11.} excerpted from the Project Gutenberg books, which contain 803M tokens and 977K unique words. 

To prepare a lexicon, we selected 522K words among the 977K unique words that occur more than once in the LibriSpeech texts. Using the base lexicon of the CMUdict\footnote{Available at http://svn.code.sf.net/p/cmusphinx/code/trunk/cmudict.} that covered 81K among the selected words of 522K, we trained a G2P model using the Sequitur tool \cite{sequitur} to cover the out-of-vocabulary words. We used the SRILM tooklit \cite{Stolcke02} to train n-gram language models (LMs). The 4-gram LM was  trained initially on the entire texts available with the modified Kneser-Ney smoothing \cite{kneserney,chen1996}, then pruned to the 3-gram LM. The first-pass decoding was conducted with the 3-gram LM and the resultant lattices were rescored with the 4-gram LM later in the second-pass. The lattices were further rescored \cite{Xu2018APR} with the neural network LMs of three TDNN layers and two LSTM layers being interleaved that were trained by the Kaldi toolkit \cite{Povey11}. 

We used the Kaldi toolkit for the acoustic modeling as well, mostly following the LibriSpeech recipe\footnote{https://github.com/kaldi-asr/kaldi/tree/master/egs/librispeech/s5.} up to the stage of Speaker Adpative Training (SAT) \cite{Gales97}. We gradually increased the training data size from 100hrs to 960hrs over the course of the GMM training stages, while the neural network training used the entire 960hr data. We first trained GMMs within the framework of 3-state Hidden Markov Models (HMMs). The conventional 39-dimensional MFCC features were spliced over 9 frames and LDA was applied to project the spliced features onto a 40-dimensional sub-space. Further projection was conducted through MLLT for better orthogonality. SAT was applied with feature-space MLLR (fMLLR) to further refine mixture parameters in GMMs. For the neural network acoustic models, we used the 40-dimensional higher-resolution MFCCs appended with 100-dimensional i-vectors \cite{Dehak11} and trained the models having the lattice alignments given by the SAT-ed GMMs as soft targets. The LF-MMI objective was used to optimized the parameters with the three regularization methods of cross-entropy, $L2$ and leaky HMM \cite{Povey16}. The exponential decrease of learning rates from $10^{-3}$ to $10^{-5}$ was applied to make the entire training procedure stable and have better convergence. The number of nodes in the final layer was determined by the number of tri-phone states in the HMM, which is 6K after the phonetic tree clustering.  The trainings were conducted on the Nvidia V100 servers with 8  GPUs.  

The dimensions of query, key,  and value embeddings in self-attention are set to $d_{q}=d_{k}=40$, and $d_{v}=80$, and $d_{model}=256$. The bottleneck dimension for the factorized convolutions and the factorized feed-forwards is 128. The number of streams and the number of the 1D convolution layers in each stream is from 1 to 5 and 0 to 7, respectively, depending on the experiment type for the ablation tests.

\begin{table}[t]
\centering
    \caption{Effect of multiple streams in WER (\%) and WERR (\%) on dev-clean and dev-other. WER: word error rate, WERR: WER reduction (relative). Lattice-rescored with the 4-gram LM. }
    \renewcommand{\arraystretch}{1.25}
    \begin{tabular}{L{2.25cm}|C{1cm}|C{1cm}|C{1cm}|C{1cm}}
        \hline
        \centering \multirow{2}{*}{\small Stream Config.} & \multicolumn{2}{c|}{\small dev-clean} & \multicolumn{2}{c}{\small dev-other}\\
        \cline{2-5} 
        \centering & \small WER & \small  WERR & \small WER & \small WERR \\
        \hline
       
        \centering \small Single (\texttt{1}) & \small 4.16 & \small  - & \small 10.66 & \small - \\
        \hline
        \centering \small \texttt{1-1-1} & \small 3.99 & \small 4.09 & \small 10.14 & \small 4.88 \\
        \hline
        \centering \small \texttt{1-1-1-1-1} & \small 3.95 & \small 5.05 & \small 10.02 & \small 6.00 \\
        \hline
    \end{tabular}
    \label{tab:adaptation}
\end{table}

\subsection{Ablation tests}
This section validates the proposed idea of multi-stream self-attention by conducting the ablation tests. 

\subsubsection{Effect of multiple streams}
As we test the validity of multiple streams in the proposed architecture in a fair comparison, we control the number of multi-heads in the self-attention layer in each stream such that $n_h^s = n_h / S$ while fixing $n_h = 15$. For example, if the number of streams is 3 (i.e., $S=3$) , then the number of multi-heads in the self-attention layer of each stream would be $15 / 3 = 5$, thus $n_h^1 = n_h^2 = n_h^3 = 5$. This is to rule out the possibility that any performance improvement would come from the significant increase of model complexity. 

Table 1 shows the effect of having multiple streams in the proposed architecture. The three entries for the system configurations in the table correspond to $S=1, 3$ and 5, respectively, while fixing the dilation rate of 1 across the streams. (For example, \texttt{1-1-1} means the three streams with the fixed dilation rate of 1 for the factorized convolutions of all the streams.) It is noticeable that the more streams we had, the better accuracy we would obtain, but without a diverse selection of dilation rates across the streams, the improvement would be limited, which will be shown more clearly in Table 2. 

\begin{table}[t]
\centering
    \caption{Effect of dilation rates in WER (\%) and WERR (\%) on dev-clean and dev-other. Lattice-rescored with the 4-gram LM.}
    \renewcommand{\arraystretch}{1.25}
    \begin{tabular}{L{2.25cm}|C{1cm}|C{1cm}|C{1cm}|C{1cm}}
        \hline
        \centering \multirow{2}{*}{\small Stream Config.} & \multicolumn{2}{c|}{\small dev-clean} & \multicolumn{2}{c}{\small dev-other}\\
        \cline{2-5} 
        \centering & \small WER & \small  WERR & \small WER & \small WERR \\
        \hline
         \centering \small Single (\texttt{1}) & \small 4.16 & \small  - & \small 10.66 & \small - \\
        \hline
        \centering \small \texttt{1-2-3} & \small 3.95 & \small 5.05 & \small 10.05 & \small 5.72 \\
        \hline
        \centering \small \texttt{1-3-5} & \small 3.99 & \small 4.09 & \small 10.27 & \small 3.66 \\
        \hline
        \hline
         \centering \small Single (\texttt{1}) & \small 4.16 & \small  - & \small 10.66 & \small - \\
        \hline
        \centering \small \texttt{1-2-3-4-5} & \small \textbf{3.84} & \small 7.69 & \small \textbf{9.94} & \small 6.75\\
        \hline
        \centering \small \texttt{1-3-5-7-9} & \small 3.85 & \small 7.45 & \small 10.03 & \small 5.91 \\
        \hline
        \centering \small \texttt{3-3-3-3-3} & \small 3.84 & \small 7.69 & \small 10.22 & \small 4.13 \\
        \hline
        \centering \small \texttt{5-5-5-5-5} & \small 3.89 & \small 6.49 & \small 10.30 & \small 3.38\\
        \hline
    \end{tabular}
    \label{tab:adaptation}
\end{table}

\subsubsection{Effect of dilation rates}
Table 2 shows the effect of having diverse dilation rates across the multiple streams of the proposed architecture. The various dilation rates across the streams are shown to help improve WERs by the clear margins. However, just mixing any values would not guarantee the performance improvement. For example, the system configuration of \texttt{1-3-5} presents the WER on the dev-other set worse than the configuration of \texttt{1-1-1} does in Table 1. It seems that a careful mix of the dilation rates would be critical for the proposed model. We found out the best configuration from the \texttt{1-2-3-4-5} setup, which has the 5 different dilation rates (by the difference of 1) for the 1D convolutions across the streams, marking 7.69\% and 6.75\% WERR on dev-clean and dev-other, respectively, as compared to the single stream baseline. This validates the efficacy of the proposed multi-stream strategy of having 1D convolutions with a unique dilation rate in each stream. The proposed architecture seemingly helps the self-attention mechanism better process embeddings in each stream and lead to more accurate results overall.

\begin{table}[t]
\centering
    \caption{Effect of factorization in the feed-forward networks of the proposed multi-stream architecture in terms of model comlexity and WER (\%) on dev-clean and dev-other. Lattice-rescored with the 4-gram LM.}
    \renewcommand{\arraystretch}{1.25}
    \begin{tabular}{C{3.4cm}|C{1cm}|C{1.25cm}|C{1.25cm}}
        \hline
        \small Stream Config. & \small Param. & \small dev-clean & \small dev-other\\
        \hline

        \centering \small Factorized-FF w/ 128-dim & \small \textbf{8M}  & \small 3.84 & \small 9.94  \\
        \hline
        \centering \small FF w/ 1,024-dim & \small 10.5M & \small 3.80 & \small 9.91 \\
        \hline
        \centering \small FF w/ 2,048-dim & \small 13M & \small 3.83 & \small 9.67 \\
        \hline
    \end{tabular}
    \label{tab:adaptation}
\end{table}

\begin{table}[t]
\centering
    \caption{Effect of 1D convolutions in WER (\%) and WERR (\%) on dev-clean and dev-other. Lattice-rescored with the 4-gram LM.}
    \renewcommand{\arraystretch}{1.25}
    \begin{tabular}{L{2.25cm}|C{1cm}|C{1cm}|C{1cm}|C{1cm}}
        \hline
        \centering \multirow{2}{*}{\small Stream Config.} & \multicolumn{2}{c|}{\small dev-clean} & \multicolumn{2}{c}{\small dev-other}\\
        \cline{2-5} 
        \centering & \small WER & \small  WERR & \small WER & \small WERR \\
        \hline
        \centering \small No Conv-F & \small 4.36 & \small  -7.13 & \small 11.86 & \small -10.43 \\
        \hline
        \centering \small 1 Conv-F & \small 4.07 & \small  - & \small 10.74 & \small - \\
        \hline
        \centering \small 3 Conv-F & \small 3.84 & \small 5.65 & \small 9.94 & \small 7.45 \\
        \hline
        \centering \small 5 Conv-F & \small 3.76 & \small 7.62 & \small 9.75 & \small 9.22 \\
        \hline
        \centering \small 7 Conv-F & \small \textbf{3.73} & \small 8.35 & \small \textbf{9.49} & \small 11.64 \\
        \hline
    \end{tabular}
    \label{tab:adaptation}
\end{table}

\subsubsection{Effect of factorized feed-forward}
The main purpose of having the factorized feed-forward networks in the proposed architecture is to retain the model complexity within a reasonable boundary even with adding more streams. The proposed multi-stream model architecture, otherwise, would increase the model complexity very easily as we add more streams. In Table 3 where we base the same configuration of \texttt{1-2-3-4-5} from Table 2, it is shown that the factorization works as expected. The factorized feed-forward networks not only contribute to the model complexity under 10M parameters but also keep the performance in a similar level with the control group that includes the normal (i.e., without factorization) feed-forward networks with the wider bottleneck layer of 1,024- or 2,048-dimension.   

\subsubsection{Effect of 1-D convolutions}
Table 4 highlights the effect of having the 1D convolutions in the proposed multi-stream self-attention model architecture. It presents the importance of the 1D convolutions preceding the self-attention mechanism in the multi-stream framework. The 7-layer Conv-F leads to roughly 15\% and 20\% WERR for the dev-clean and dev-other dataset, respectively, against the case of having no convolutions. The pattern seems that the more convolution layers, the more performance gain would be obtained, but after the 7 layers we didn't observe any significant performance boost. 

\begin{table}[t]
\centering
    \caption{Comparison of the best configured model in terms of LMs in WER (\%). 3-T/2-L: 3-layer TDNNs \& 2-layer LSTMs being interleaved. The numbers in the parentheses indicate the size of neurons.}
    \renewcommand{\arraystretch}{1.25}
    \begin{tabular}{L{2.5cm}|C{1cm}|C{1cm}|C{1cm}|C{1cm}}
        \hline
        \centering \multirow{2}{*}{\small System} & \multicolumn{2}{c|}{\small dev} & \multicolumn{2}{c}{\small test}\\
        \cline{2-5} 
        \centering & \small clean &  \small  other & \small clean & \small other \\
        \hline
        \centering \small 4-gram & \small 2.65 & \small 8.21 & \small 2.93 & \small 8.32 \\
        \hline
         \centering \small 3-T/2-L (1,024) & \small 2.51 & \small 7.13 & \small 2.69  & \small 7.45 \\
        \hline
        \centering \small 3-T/2-L (2,048) & \small 2.23 & \small 6.82 & \small 2.39  & \small 7.01 \\
        \hline
        \centering \small 3-T/2-L (4,096) & \small 2.14 & \small 6.51 & \small 2.21 & \small 6.73 \\
        \hline
        \centering \small 4-LSTM (2,048) & \small 1.84 & \small 5.75 & \small 2.20 & \small 5.82 \\
        \hline
    \end{tabular}
    \label{tab:adaptation}
\end{table}

\subsection{Comparison with the other state-of-the-arts}
In this section, we configure our best model for the LibriSpeech speech recognition task by stacking the proposed multi-stream self-attention blocks. The chosen configuration of the best ASR system for us is
\begin{itemize}
    \item 3 layers of multi-stream self-attention blocks
    \item 5 streams of self-attention encoders in each block
    \item 7 layers of 1D Conv-F's in each stream
    \item Dilation configuration of \texttt{1-2-3-4-5} to 1D CONV-F's across streams 
\end{itemize}
The total number of parameters for this setup is 23M. The lattice-rescored result of this system with the 4-gram LM is presented in the first row in Table 5.

As for the neural network language models, we trained the models of 3 TDNN layers and 2 LSTM layers (uni-directional) being interleaved. The averaged (relative) error rate reduction by the best neural network language model (with the dimention of 4,096) against the 4-gram LM case ranges from 15\% to 20\%. The embedding dimension seems to matter in terms of performance improvement. We did not try the bigger dimensions than 4,096 due to the prohibitive model training time. The best model in terms of WER improvement is the one with the 4-layered LSTMs, trained with around 10K word pieces. 

Table 6 shows some state-of-the-art system performances. The hybrid DNN/HMM approach was used for the different neural network acoustic models of CNN-biLSTM, pyramidal Feedforward Sequential Memory Network (FSMN), and BiLSTM, respectively, in \cite{capio,cloudwalk,rwth}. The Transformer LM was exploited as the best rescoring LM in \cite{rwth}. In \cite{tds,specaugment} the end-to-end framework was considered. Time Depth Separable (TDS) convolutions were introduced in \cite{tds} while the cut out of spectrograms was applied on the fly during training to enhance the noise robustness. In \cite{karita}, the full Transformer model was studied comparatively in various speech tasks. As compared to the other state-of-the-art system performances, it is shown that we achieve the best performances on both of the dev-clean and test-clean set, while on the dev-other and test-other set the lowest WERs are presented in \cite{rwth}. The WERs of 1.8\% and 2.2\% on the datasets by the proposed system are the best numbers thus far reported in the literature.

\begin{table}[t]
\centering
    \caption{Comparison with the other state-of-the-art systems in WER (\%).}
    \renewcommand{\arraystretch}{1.25}
    \begin{tabular}{L{2.5cm}|C{1cm}|C{1cm}|C{1cm}|C{1cm}}
        \hline
        \centering \multirow{2}{*}{\small System} & \multicolumn{2}{c|}{\small dev} & \multicolumn{2}{c}{\small test}\\
        \cline{2-5} 
        \centering & \small clean & \small  other & \small clean & \small other \\
        \hline
        \centering \small Han, et al. \cite{capio} & \small 3.1  & \small 8.3 & \small 3.5 & \small 8.6 \\
        \hline
        \centering \small Hannun, et al. \cite{tds} & \small 3.0 & \small 8.9 & \small 3.3 & \small 9.8 \\
        \hline
        \centering \small Yang, et al. \cite{cloudwalk} & \small 2.6 & \small 7.5 & \small 3.0 & \small 7.5 \\
        \hline
        \centering \small Karita, et al. \cite{karita} & \small 2.2 & \small 5.6 & \small 2.6 & \small 5.7 \\
        \hline
        \centering \small Park, et al. \cite{specaugment} & \small - & \small - & \small 2.5 & \small 5.8 \\
        \hline
        \centering \small L{\"u}scher, et al. \cite{rwth} & \small 1.9 & \small \textbf{4.5} & \small 2.3 & \small \textbf{5.0} \\
        \hline
        \centering \small Proposed & \small \textbf{1.8} & \small 5.8 & \small \textbf{2.2} & \small 5.8  \\
        \hline
    \end{tabular}
    \label{tab:adaptation}
\end{table}

\section{Conclusions}

We proposed the multi-stream self-attention model architecture preceded by the layers of the factorized 1D convolutions with the unique dilation rate in each stream. This architecture allows input speech frames to be efficiently processed and effectively self-attended. We validated the proposed ideas by performing the ablation tests, and also configured the state-of-the-art ASR system by stacking the multi-stream self-attention model blocks with the strong neural network language models. The WERs on the dev-clean and test-clean set of 1.8\% and 2.2\% are the best reported numbers found in the literature.

Note that the proposed system has only 23M parameters. The other systems in Table 6 have much higher model complexity, for example,  100M for CNN-biLSTM \cite{capio} or 200M for LAS in \cite{specaugment}. This could make our system more practical and appealing to speech engineers or practitioners who would like to deploy ASR models as a service on devices with the limited computing power. Of course, the practicality of the proposed model on on-device ASR would rely upon the usage of much lighter LM models. 

We plan to further explore the self-attention mechanism to make it more suitable for speech applications. To enhance the noise robustness especially on the dev-other and test-other data of the LibriSpeech corpus, we will continue to work on the data augmentation side. In this paper, we used speed perturbation as the data augmentation strategy and it is worth considering other types of augmentation such as data dropout like the cut out method used in \cite{specaugment}.


\bibliographystyle{IEEEbib}
\bibliography{refs}

\end{document}